\documentclass[conference]{IEEEtran}
\IEEEoverridecommandlockouts
\usepackage{cite}
\usepackage{amsmath,amssymb,amsfonts}
\usepackage{algorithmic}
\usepackage{graphicx}
\usepackage{textcomp}
\usepackage{xcolor}
\usepackage{float}
\usepackage[caption = false]{subfig}
\def\BibTeX{{\rm B\kern-.05em{\sc i\kern-.025em b}\kern-.08em
    T\kern-.1667em\lower.7ex\hbox{E}\kern-.125emX}}
\begin{document}

\title{Characterization and recognition of handwritten digits using Julia\\
}

\author{\IEEEauthorblockN{M. A. Jishan $^{1}$, M. S. Alam $^{2}$, Afrida Islam $^{3}$, I. R. Mazumder $^{4}$, K. R. Mahmud $^{5}$ and A. K. Al Azad $^{6}$}

\IEEEauthorblockA{Faculty of Statistics, Technische Universität Dortmund, Germany $^{1}$, $^{2}$, $^{3}$, $^{4}$}
\IEEEauthorblockA{Department of Computer Science and Engineering, University of Liberal Arts Bangladesh $^{5}$, $^{6}$}
{\{md-asifuzzaman.jishan, md-shahabub.alam, afrida.islam, imran.mazumder\}@tu-dortmund.de,\{raqib.mahmud, abul.azad\}@ulab.edu.bd}}

\maketitle
\begin{abstract}
Automatic image and digit recognition is a computationally challenging task for the image processing and  pattern recognition, requiring an adequate appreciation of the syntactic and semantic importance of the image for the identification of the handwritten digits. Image and Pattern Recognition has been identified as one of the driving forces in the research areas because of its shifting of different types of applications, such as safety frameworks, clinical frameworks, diversion, and so on. In this study, for recognition, we implemented a hybrid neural network model that is capable of recognizing the digit of MNIST dataset and achieved a remarkable result. The proposed neural model network can extract features from the image and recognize the features in the layer by layer. To expand, it is so important for the neural network to recognize how the proposed model can work in each layer, how it can generate output, and so on. Besides, it also can recognize the auto-encoding system and the variational auto-encoding system of the MNIST dataset. This study will explore those issues that are discussed above, and the explanation for them, and how this phenomenon can be overcome.
\end{abstract}


\section{Introduction}
The importance of good metrics and simplified problems cannot be downplayed. In particular, in critical and fast-moving fields such as artificial intelligence, computer vision and pattern recognition. These tasks offer a simple, quantitative, and rational way to break down and analyze the evolution of learning methods and strategies. In particular, where the errand is instinctive and addictively quick, observers will quickly understand the layout and characteristics of the methods and calculations.

As a single dataset may just cover a particular or specific task, the presence of a fluctuated set-up of benchmark errands is significant in permitting a more comprehensive way to deal with evaluating and portraying the exhibition of a calculation or framework. In the machine learning sector, there are a few normalized datasets that are broadly utilized and have gotten profoundly serious. These incorporate the MNIST dataset [1], the CIFAR-10 and CIFAR-100 datasets, the STL-10 dataset, and Street View House Numbers (SVHN) dataset [2], [3], [4], [5], [6], [7].

The MNIST dataset is the most widely recognized and used dataset for image recognition and computer vision. It includes a 10-class hand-written mission which was first published in 1998. In order both to provide assistance and to guarantee its lifetime, a decent dataset is required to address an adequately moving issue. This is perhaps where MNIST has endured the use of deep learning and convolution neural systems despite their infinitely high precision. Different exploration bunches distributed more than 99.7 percent accuracy. It is an order accuracy in which the name of the dataset may raise questions. There is preciseness in order that questions can be posed about the name of the dataset [5], [6], [7], [8], [9].

Image processing refers to the mastery that recognizes locations, objects, artifacts, structures, faces, and some other object factors. We exchange huge measurements of knowledge with software, organizations, and websites. Besides, cell phones with cameras produce limitless computerized images and graphics [10], [11], [12]. The vast quantity of skills is used to communicate with the people that reach them better and more intelligent management. The key steps in the image recognition technology are data compilation and sorting, predictive model creation, and photo identification. These models match the capacity of the human visual system (HVS). The human eyes see an image as several colors produced by the visual cortex in the cerebrum. These results, far away from the scene, relate to ideas and articles recorded in one's memory. Image recognition tries to emulate this process. 
Computer considers the image to be a group of pixels with discreet numbering colors (Red, Green, and Blue). Visual perception is a big part of the vision of the computer. The main aim of AI is to produce input information calculations that benefit from the factual investigation to predict appropriate output estimates. The main aim of AI innovation is the machine receives all the information and uses it to justify when facing a different configuration. In collaboration with the planet, machines constantly learn and develop their knowledge collection. The concept of machine learning is frequently found at the core of these systems for learning and transmission with the extension of the wise systems.

The Julia programming language is designed specifically for scientific computing. It is a flexible dynamic language with performance comparable to traditional static-typed languages. Julia aims to provide a single environment that is productive enough for prototyping and efficient for industrial applications. It has a very intuitive model that suits what is expected from a machine learning programming language. We implemented the Julia programming language and their machine learning library for the characterization and recognition of the MNIST handwritten dataset.

The research investigates the filters which are model parameters redundancy by visualization method. The key point of the research is using a target dataset e.g. MNIST implemented by Julia and our system follows the conventional procedure of extracting features from an image using a simple Convolutional Neural Network. Moreover, we represented thresholds value, the filter number, ratio of similar pairs, visual of similarities, a testing input image, and output using activation function during the similar filters in the first convolutional layer. We have used the Cosine Similarity algorithm to work out the similarity between two filters which is responsible for detecting similar filters. Furthermore, we showed that result for similar filters in the second convolutional layer with filter size $3\times3$ with using the filter size from 32 to 256. Besides, we also represented similar filters in the second convolutional layer using filter size using $5\times5$. Finally, we represented autoencoding and variational auto-encoder results using MNIST. By visualizing the activation maps of the filters, we have validated the system for determination.

\section{Recent Work}
The paper “Characterization of Symbolic Rules Embedded in Deep DIMLP Networks: A
Challenge to Transparency of Deep Learning complies rule extraction from ensembles of Discretized Interpretable Multi Layer Perceptrons (DIMLP) and deep DIMLPs for predictive accuracy on digit recognition. Feature detectors created by neural networks over the MNIST as
well as the complexity of the extracted rulesets is possibly determined to keep well balanced
between accuracy and interpretability [13], [14]. Whereas, a way of reducing the number of parameters
in fully connected layers of a neural network using pre-defined sparsity can be derived. That
indicated convolutional neural networks can operate without any loss of accuracy at less than
0.5-5 percent overall network connection density [15], [16], [17]. The results can be shown in MNIST.

Restricted Boltzmann Machines (RBM) can generate graded and distributed representations of
data. The work- “Emergence of Compositional Representations in Restricted Boltzmann
Machines” has shown, how characterizing the structural conditions and allowing RBM to operate
in such compositional phase, RBM is trained on the handwritten digits of MNIST [18]. Moreover,
as the Neural Networks structure is inherently computing and power intensive, so hardware
accelerators emerge as a promising solution. Through a High Level Synthesis (HLS) approach,
characterizing the vulnerability of several components of Register-Transfer Level (RTL) model
of Neural Network it is shown that severity of faults depends on application level specification
and Neural Network Data [19], [20].

By using response characterization methods, a systematic pipeline for interpreting individual
hidden state dynamics within Recurrent Neural Networks (RNNs), especially Long Short-term
Memory Networks (LSTMs) can be defined at the cellular level. This method can uniquely
identify neurons with insightful dynamics and test accuracy through ablation analysis [20].
Another method is proposed for discovering features required for separation of images using
deep autoencoder. In this certain methodology, it auto learns the image representation features
for clustering and groups the similar images in a cluster simultaneously separating the dissimilar
images into another cluster [18, [19], [20], [21].

A comparison of four neural networks on MNIST Dataset: Convolutional Neural Networks
(CNN), Deeps Residual Network (ResNet), Dense Convolutional Network (DenseNet) and
improvement on CNN baseline thus Capsule Network (CapsNet) for image recognition is also
done. The CapsNet is considered as giving excellent performance despite having small amount
of data [22]. However, “EMNIST: an extension of MNIST to handwritten letters” make
alterations to the MNIST dataset named as Extended MNIST constituting more challenging
dataset allowing for direct compatibility with all existing classifiers and systems [22], [23].

\section{Dataset}

The neural system works on the feeding dataset. The MNIST (Modified
National Institute of Standard and Technology database) is an
open source handwritten digits dataset, widely used for training and
testing in the field of machine learning, computer vision, and image
processing. This work is done using the MNIST dataset, feeding into
the neural network system, directly imported and downloaded from
Keras. That dataset containing 60,000 training and 10,000
testing images are stored in a simple file format design for sorting
vectors as well as multidimensional matrices. The database is a subset of
the samples of black and white images of NIST’s original datasets,
which data were collected by Lecun et al from United States Census
Bureau employees and high school students. In the MNIST, the
handwritten digits are size-normalized and centered in a fixed-size $28\times28$ images with corresponding labels, split into a training, validation, and
test set. The training set shows the neural system gauging various
highlights, whereas to predict the response validating set is used. The
testing set helps in providing an impartial result of the final model fitting
into the training dataset. To prepare the model to arrange these pictures into the right class (0-9) is the target of this work [3], [4], [5].

\section{Methodology}

In the area of Computer Vision, a neural system framework requires complex computation that empowers the computational framework to discover designs by coordinating complex info information connections like human minds.

Convolutional Neural Network (CNN) is a technique of deep neural network model that chooses attributes in the taken picture and separates from others. In earlier years, channels like a blur, sharpening, and identifying edge was should have been hand-designed and included enough preparation before CNN becomes an integral factor. The wide execution of this calculation, for example, has culminated in facial expression recognition, picture allocation and object identification, proposal framework, handwritten recognition, and image to natural language processing system [24].

\begin{figure}[htbp]
\centerline{\includegraphics[scale=0.56]{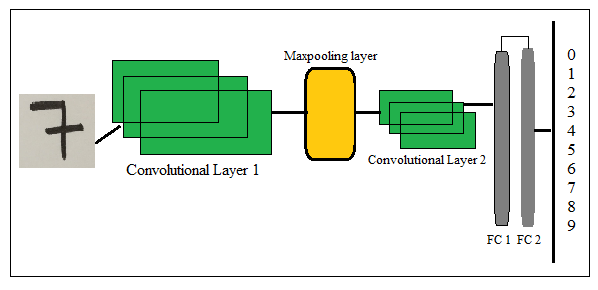}}
\caption{Image characterization result of CNN part.}
\label{fig}
\end{figure}

We utilized four primary layers of the design of CNN: Convolutional Layer, Pooling Layer, Rectified non-linear unit, and Fully-Connected Layer. Convolution layer is available at the focal point of the system and performs convolutions that include straight activity using augmentation, a lot of loads with a variety of info information called to channel or piece. The fundamental reason for convolution is to bring top-notch highlights like edge location and some inferior quality highlights like shading, slope direction, and so on [24], [25], [26]. Utilizing a similar channel to identify a specific article in the picture has been perceived amazing as it will sift through methodically everywhere throughout the picture where the item is available [27], [28].

The following comes the pooling layer. The primary goal is to constantly diminish the spatial size of the portrayal to diminish the number of boundaries and calculation in the system just as controlling overfitting [29], [30]. With the assistance of MAX activity, it works freely to each profundity cut of the info and resizes it spatially. Completely associated layer: Neurons in an FC layer have full associations with all initiations in the past layer, their enactments can thus be registered with a network increase followed by an inclination counterbalance. It is conceivable to change over FC layers to CONV layers as there is extremely little contrast. We represented the proposed model in Figure 1.

\section{\uppercase{Simulation Setup}}
\label{sec:introduction}

\subsection{Image Handling}

We implemented the convolutional neural network for our model. That CNN Network model comprises of two convolutional neural network layers and one Maxpooling2D layer that down samples the output from the convolutional blocks. We utilized the filter size of $3\times3$ for the first convolutional layer and after that, we implemented $5\times5$ for the second convolutional layer. The input image of the model is of shape $28\times28$ with one color channel.

We utilized crude picture documents of dataset close by CNN and VGG highlights. We set pixels measure $224\times224$. The pictures of the MNIST dataset are without question concealing pictures with pixel regards running from 0 to 255 with a segment of $224\times224$, so before feed the data into the model, it is indispensable to pre-process it. First devotee each $224\times224$ picture of the dataset into a matrix of size $224\times224$, which we would then be able to take care into the CNN implementation. We focused on the Julia programming language for implementing this MNIST dataset and use flux, plots, statistics, mnist, onehotbatch, onecold, crossentropy, throttle, repeated packages for multilayer perception section. During auto encoding, we use flux, mnist, epochs, onehotbatch, mse, plots, throttle, partition, and juno packages. For variational autoencoding, we use flux, mnist, throttle, params, juno, plots, distributions, and epochs for implementation.

This requires the picture to be gone through a few preprocessing capacities that convert the image to a worthy shape. After the last convolutional layer, there is a flattened layer, which has 1936 neurons. Our CNN is as yet handling matrix and we have to change over those units into the vector to take care of into the completely associated arrange for the ultimate result. So, we apply a smooth layer here. It changes over the entire systems loads to vector from the matrix. In any case, toward the finish of the system, there are 2 completely associated layers exist. The principal completely associated layer has 128 neurons, the subsequent one has 50 neurons. Finally, we utilized the softmax classifier technique for ordering the output, as the dataset has 10 distinct classes so characterizes the yield into 10 neurons.

We implemented categorical \texttt{\char`_}crossentropy as loss function to measure how good our model is, or how our model works and gives output in perfectly. For classification, we have used a softmax classifier. It is placed at the output layer of a neural network. It is commonly used in multi-class learning problems. Softmax function takes an N-dimensional vector of real numbers and transforms it into a vector of real number in range (0, 1) which adds up to 1.

\subsection{Technique of the Optimization}
For the improvement territory of CNN, we utilized a rectified linear unit (ReLU) analyzer for the MNIST dataset using Julia. We utilized learning rate 0.001, rot rate=1e-6, momentum=0.9, nesterov=True. We cross-endorse the learning rate and weight decay. We used dropout regularization techniques in all layers. We utilized ReLU and softmax initiation work close by set the dropout layer and utilized the distinctive edge esteems for the diverse convolutional layers. We also maintain accuracy and loss in the CNN part for measurement accuracy and loss vs. epoch.

\section{Results and Discussion}
This research focused on the handwritten recognition using the MNIST and Julia programming language and also focused on the different type of filters in neural networks along with fine-tuning it. This research also concentrated on which channels in neural systems might be excess, on the off chance that they have insignificantly little qualities if their usefulness is imitated by another channel while expanding or diminishing the number of channels and comparability rate in various convolutional layers. During the training period, in this research train up the full dataset using 50 epochs and achieved 0.991926 for the training time accuracy. Additionally, within Figure 2 and 3 training time accuracy and loss are represented graphically. In this research, also implemented the Threshold, which will determine the range, for this, we have organized the system with different setups.

\begin{figure}[htbp]
\centerline{\includegraphics[scale=0.66]{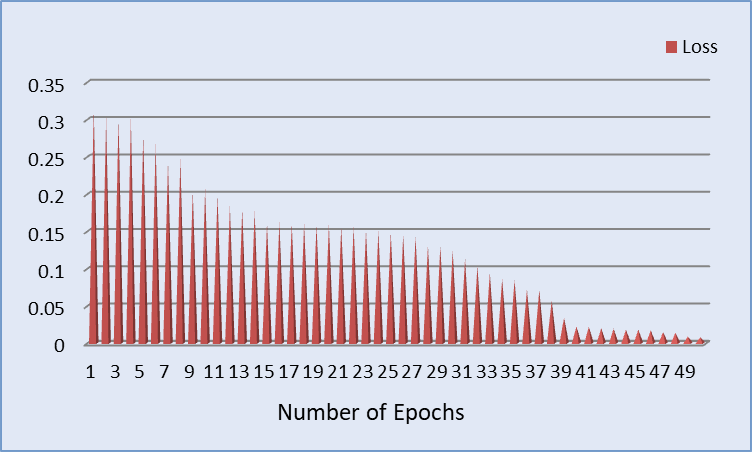}}
\caption{Graphical representation of during training time loss.}
\label{fig}
\end{figure}

\begin{figure}[htbp]
\centerline{\includegraphics[scale=0.66]{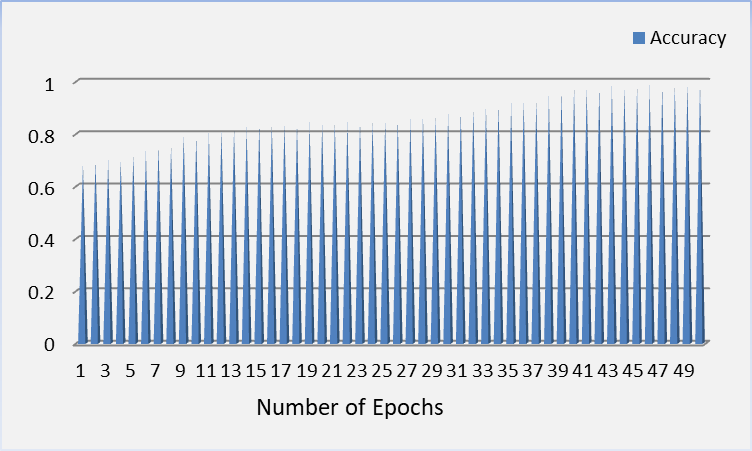}}
\caption{Graphical representation of during training time accuracy.}
\label{fig}
\end{figure}

In the wake of building up a framework, we took care of the dataset into the framework and prepared the up to 50 epochs. Accept that in the first layer the filter size was $3\times3$ and the number of the filter is 256 and in the second convolutional layer our channel size is $5\times5$ and the number of filters size was 256 along with implemented batch size 128. This research achieved 0.981411 for validation period accuracy results and represented graphically in Figure 4 and 5 along with validation period loss. 

\begin{figure}[htbp]
\centerline{\includegraphics[scale=0.38]{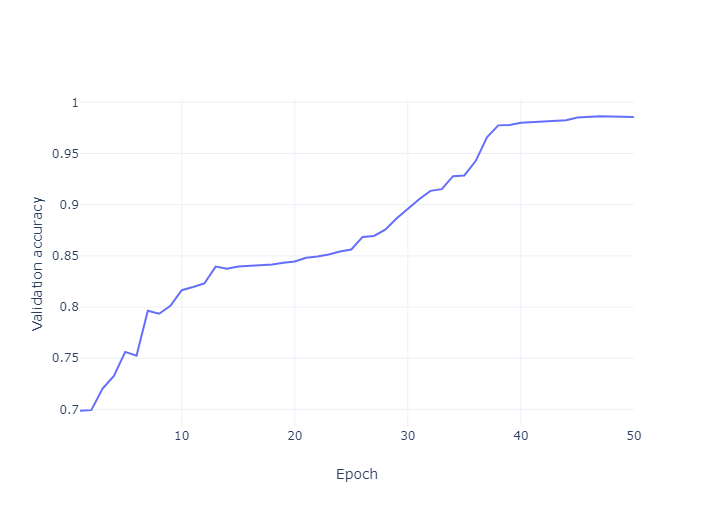}}
\caption{Graphical representation of validation time accuracy.}
\label{fig}
\end{figure}

\begin{figure}[htbp]
\centerline{\includegraphics[scale=0.38]{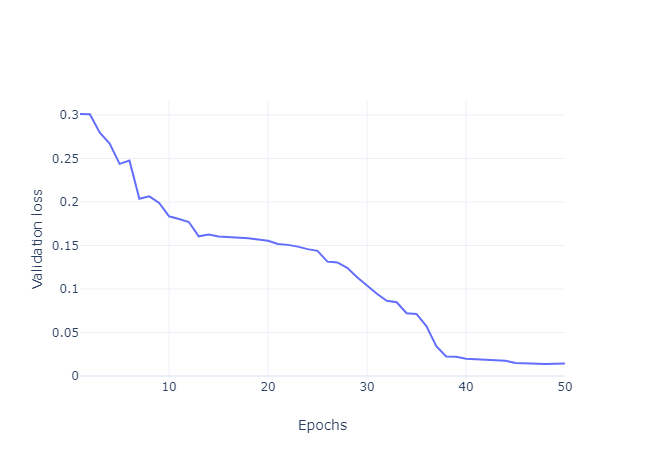}}
\caption{Graphical representation of validation time loss.}
\label{fig}
\end{figure}

The model predicts the comparable pair of filters on given filter size and the number of filters is utilizing cosine closeness. First, we have prepared our framework with the MNIST dataset using Julia and anticipated that if the framework is learning the same highlights over and over. To extend, we have examined a few tests to assess our framework and legitimize our exploration. We have taken our convolutional layer with a filter size of $3\times3$ along with $5\times5$ and the number of channels is 32 to 256 with two unique thresholds value which worth are 0.5 and 0.6 respectively. We have removed just one set of comparable channels and envisioned it. Moreover, we represented thresholds value, the filter number, ratio of similar pairs, visual of similarities, a testing input image, and output using activation function during the similar filters in the first convolutional layer in Figure 7. Furthermore, within Figure 8, we showed that result for similar filters in the second convolutional layer with filter size $3\times3$ with using the filter size from 32 to 256. In addition, we also represented similar filters in the second convolutional layer using filter size $5\times5$ within Figure 9.

\subsection{Auto-encoder}
An auto-encoder is an unsupervised method for neural network learning which reduces data size by learning to ignore data noise. Effective data representation (encoding) is often taught through network training to suppress "dust" signals. For the show auto-encoding, the info measurement is ${(28)}^2$ and the output of measurement of the encoder is 32. We demonstrated our auto-encoding output in Figure 6.

\begin{figure}[htbp]
\centerline{\includegraphics[scale=0.43]{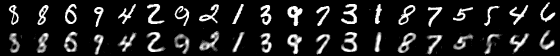}}
\caption{Output of auto encoding system for MNIST dataset.}
\label{fig}
\end{figure}

\begin{figure}[htbp]
\centerline{\includegraphics[scale=0.55]{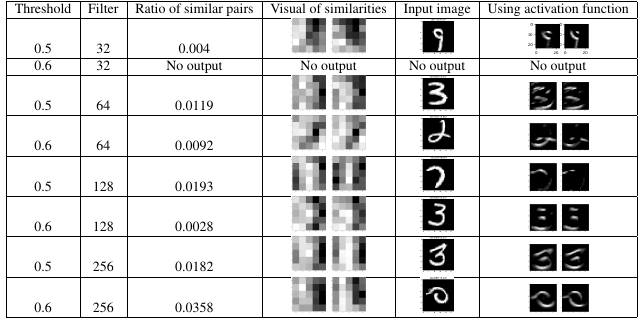}}
\caption{Similar filters of the first convolutional layer.}
\label{fig}
\end{figure}

\begin{figure}[htbp]
\centerline{\includegraphics[scale=0.55]{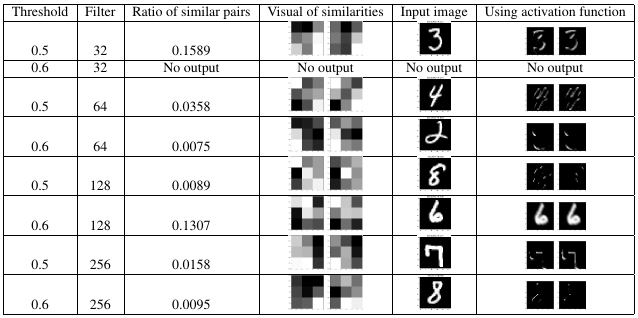}}
\caption{Similar filters of the second convolutional layer with filter size $3\times3$.}
\label{fig}
\end{figure}

\begin{figure}[htbp]
\centerline{\includegraphics[scale=0.55]{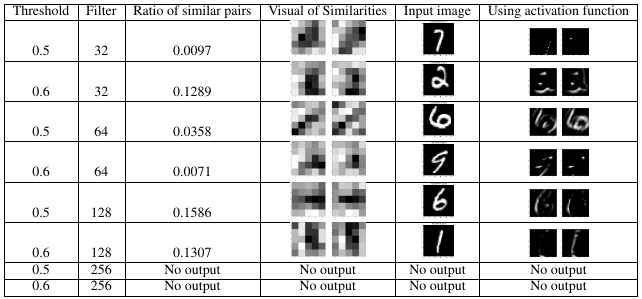}}
\caption{Similar filters of the second convolutional layer with filter size $5\times5$.}
\label{fig}
\end{figure}

\subsection{Variational Autoencoder }

In neural network framework language, a VAE comprises an encoder, a decoder, and a deficit work. In likelihood model terms, the variational autoencoder alludes to inexact inference in an idle Gaussian model where the rough back and model probability is parameterized by neural nets. In Figure 10, we demonstrated the variational autoencoder results for our research which was implemented by MNIST and Julia.

\begin{figure}[htbp]
\centerline{\includegraphics[scale=0.79]{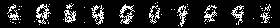}}
\caption{Output of Variational auto encoding system for MNIST dataset.}
\label{fig}
\end{figure}

\subsection{Discussion }
We illustrated a CNN model that is capable of characterization of a handwritten digit dataset using Julia. That model categorized the MNIST dataset and we represented graphically the training time accuracy which is 0.991926 and loss results in Figure 2 and 3. That model categorized the MNIST dataset and we represented graphically the training time accuracy which is 0.991926 and loss results in Figure 2 along with Figure 3. After that, we also illustrated the validation time result which is 0.981411 in Figure 4 and 5. Moreover, we showed the thresholds value, the filter number, the ratio of similar pairs, visual of similarities, a testing input image, and output using activation function during the similar filters in the first convolutional layer in Figure 7. 

Furthermore, within Figure 8, we showed that result for similar filters in the second convolutional layer with filter size $3\times3$ with using the filter size from 32 to 256. To extend, we also represented similar filters in the second convolutional layer using filter size $5\times5$ within Figure 9. Finally, we demonstrated our auto-encoding and the variational autoencoder results in Figures 6 and 10.

\section{Conclusion}
In this study, we propose a neural network model that is capable of characterization and recognition of the MNIST dataset and identify the handwritten object using Julia. Our proposed model achieved the best accuracy during the training and testing time. Moreover, this research has also been concerned about implemented the Threshold, which will determine the range, for this, we have organized the system with different setups. After that, in the first layer of model the filter size was $3\times3$ and the number of the filter is 256 and in the second convolutional layer our channel size is $5\times5$ and the number of filters size was 256 along with implemented batch size 128. To extend, we have taken our convolutional layer with a filter size of $3\times3$ along with $5\times5$, and the number of channels is 32 to 256 with two unique thresholds value which worth is 0.5 and 0.6, respectively, and our study showed that how one can change the image size and visualization. After that, we also represented the auto- encoding system result, and the variational autoencoder result using Julia. Moreover, we have an intension in the future to improve accuracy by implementing using Julia an extraordinary characterization and recognition of handwritten dataset.

\end{document}